\documentclass[11pt,a4paper]{article}
\usepackage[hyperref]{acl2021}
\usepackage{times}
\usepackage{latexsym}

\usepackage{microtype}
\usepackage{enumitem}
\usepackage{bookmark}
\usepackage{bm}
\usepackage{amsfonts}
\usepackage{subfigure}
\usepackage{graphicx}
\usepackage{booktabs}
\usepackage{makecell}
\aclfinalcopy 

\usepackage{color}

\title{REAM$\sharp$: An Enhancement Approach to Reference-based Evaluation Metrics for Open-domain Dialog Generation}

\author{Jun Gao \\
  \normalsize Tencent AI Lab, Shenzhen, China \\
  \normalsize \texttt{imgaojun@gmail.com} \\\And
  Wei Bi\thanks{~~Corresponding authors} \\
  \normalsize Tencent AI Lab, Shenzhen, China \\
  \normalsize \texttt{victoriabi@tencent.com} \\\AND
  Ruifeng Xu$^*$ \\
  \normalsize Peng Cheng Laboratory, Shenzhen, China \\
  \normalsize  \texttt{xuruifeng.hitsz@gmail.com} \\\And
   Shuming Shi \\
  \normalsize Tencent AI Lab, Shenzhen, China \\
  \normalsize  \texttt{shumingshi@tencent.com}
  }

\date{}

\begin{document}
\maketitle
\begin{abstract}
    The lack of reliable automatic evaluation metrics is a major impediment to the development of open-domain dialogue systems.
    Various reference-based metrics have been proposed to calculate a score between a predicted response and a small set of references. However, these metrics show unsatisfactory correlations with human judgments. For a reference-based metric, its reliability mainly depends on two factors: its ability to measure the similarity between the predicted response and the reference response, as well as the reliability of the given reference set. Yet, there are few discussions on the latter. Our work attempts to fill this vacancy. We first clarify an assumption on reference-based metrics that, if more high-quality references are added into the reference set, the reliability of the metric will increase. Next, we present \textbf{REAM$\sharp$}: an enhancement approach to \textbf{R}eference-based \textbf{E}v\textbf{A}luation \textbf{M}etrics\footnote{Interested reader may contact the authors to obtain a copy of the code and the data.} for open-domain dialogue systems. A prediction model is designed to estimate the reliability of the given reference set. We show how its predicted results can be helpful to augment the reference set, and thus improve the reliability of the metric. Experiments validate both the effectiveness of our prediction model and that the reliability of reference-based metrics improves with the augmented reference sets.
  \end{abstract}
  
  \section{Introduction}
  
  
  The lack of reliable automatic evaluation metrics is a major impediment to the development of open-domain dialogue systems~\citep{Li2016MutualIA,Gao2019GeneratingMD,Li2020RelevancePromotingLM}. The underlying difficulty in evaluation lies in the diversity of the possible outcomings. 
  Existing evaluation metrics for open-domain dialogue systems can be roughly divided into reference-based and reference-free metrics. Reference-based metrics usually measure how similar a generated response is to the reference responses.  Reference-free metrics, on the other hand, measure the quality of a response without any reference and usually focus on
specific aspects of the responses. For example, much work often computes the perplexity of a generated response as a measure of fluency~\citep{li2020slot}, and adopts Dist-1/2~\citep{Li2016ADO} to measure the diversity of the response. In this work, we focus on reference-based metrics.
  
  

  
  

  BLEU~\citep{Papineni2002BleuAM}, originally for machine translation, is now a popular reference-based metric to evaluate open-domain dialog systems  automatically. 
  However, it has been shown that BLEU and other word-overlap metrics such as METEOR~\citep{Banerjee2005METEORAA} and ROUGE~\citep{Lin2004ROUGEAP}, rely on surface-form similarities only without considering the semantic diversity, thus fail to correlate well with human judgements~\citep{Liu2016HowNT}. 
  Instead, embedding-based metrics are adopted to consider the semantic meaning of a word defined by a distributed representation. 
  For example, \citet{Zhang2020BERTScoreET} introduce an embedding-based metric BERTScore that computes the similarity between the generated response and reference responses using contextual embeddings obtained from BERT~\citep{Devlin2019BERTPO}.
  Intuitively, the reliability of a referenced-based metric depends on two factors: (1) the ability of the metric to measure the similarity between the generated response and the reference response and (2) the reliability of the reference set for evaluating each generated response.
  As can be seen from above, most current work falls into improving the former factor, while few considers the latter.
  However, without a high-quality reference set, the results obtained by all these metrics will have a poor correlation with human judgments. 
  
  Unlike most previous studies that propose new evaluation metrics, the focus of our study is on improving the reliability of the reference set.
  We first clarify an assumption on reference-based metrics that, if more high-quality responses are added into the reference set, the correlation of reference-based metrics with human judgments will increase. 
  We perform experiments to demonstrate that
  the standard BLEU~\cite{Papineni2002BleuAM} does not hold this assumption, but two existing metrics can.
  One is a modified BLEU metric~\cite{Freitag2020BLEUMB} that compares the generated response with each reference response within the set using single-reference BLEU. We refer this modified BLEU as BLEU*. 
  The other is the BERTScore~\cite{Zhang2020BERTScoreET}, which can also be used to evaluate responses with multiple references. 

 
  In this work, we propose \textbf{REAM$\sharp$}: an enhancement approach to \textbf{R}eference-based \textbf{E}v\textbf{A}luation \textbf{M}etrics for open-domain dialogue systems.
  Our approach, which can enhance a reference-based metric that satisfies our assumption (such as BLEU* and BERTScore), consists of two parts: (1) reliability prediction and (2) high-quality references augmentation. 
  In the first part, we devise a reliability prediction model to estimate the reliability of the reference set. Given a query and its reference set, the model will predict a reliability score to reflect how reliable a metric is used to evaluate the results of the query using the given reference set.
  In the second part, we aim to augment high-quality references with the help of the reliability prediction model. We introduce   two ways to handle reference candidates with different qualities. If the acquired reference candidates are considered reliable, we can adopt automatic annotation. If we are not certain about the relevance of the reference candidates, human annotators are needed for an interactive annotation.
    Experimental results show that our proposed approach can effectively enhance the reliability of the reference set and improve the correlation of the reference-based metrics BLEU* and BERTScore.
    
    The rest of this paper is organized as follows. In Section~\ref{sec:sec2}, we introduce our assumption on the reference-based metrics, and conduct a series of preliminary experiments to validate this assumption on existing metrics. We also provides details about data collection and metric evaluation in this section. Section~\ref{sec:approach} describes our reliability prediction model and Section~\ref{sec:augment} presents how to augment high-quality references with the help of the proposed reliability prediction model. Section~\ref{sec:main_exp} shows our experimental results about the proposed reliability prediction model as well as different strategies to augment the reference set. Section~\ref{sec:related_work} describes related work. Finally, we conclude our 
    work and discuss some future work in Section~\ref{sec:conclusion}. 

  \section{Research Questions and Settings}
  \label{sec:sec2}
  \label{sec:assumption}
  We first make an assumption on reference-based metrics for open-domain dialog system that if more high-quality reference responses are added to the reference set, a reference-based metric will show a higher correlation with human judgments. 
  We want to draw such an assumption due to two considerations.
  First, by considering the nature of open-domain dialog, a query is possible to be relevant to multiple diverse responses.
  Including more relevant responses in the reference set can naturally help alleviate the assessment difficulty associated with linguistic variation.
  Second,
  if a low-quality response, e.g. a general response ``I don't know'' is added to the reference set, the reference-based metric will assign a very high score for the same low-quality predicted response, resulting in a low correlation. Therefore, only by ensuring that the responses in the reference set are of high-quality can we avoid the metric assigning high scores to low-quality responses.
  
  We validate this assumption on three existing metrics: the original multi-reference BLEU~\cite{Papineni2002BleuAM}, BLEU*~\citep{Freitag2020BLEUMB} and BERTScore~\citep{Zhang2020BERTScoreET}.
  The original BLEU uses a modified form of precision to compare a generated response against multiple reference responses. For BLEU* and BERTScore, the multi-reference score of a response $y$ can be computed as: 
  \begin{equation}
      score(y,R)=\mathop{\mathrm{max}}\limits_{r\in R}d(y,r)
  \end{equation}
  where $d$ is the given metric, $R=\{r_1,r_2,\cdots,r_n\}$ is the given reference set.
  
  We examine the assumption by answering the following questions:
  \begin{enumerate}[noitemsep, topsep=0pt]
    \item  \textit{Will the correlation of the reference-based metric improve when more high-quality responses are added to the reference set?}
    \item \textit{How will the low-quality responses included in the reference set affect the correlation of the reference-based metric?}
    
  \end{enumerate}
  \noindent
  Before presenting the details of the preliminary experiments, we first describe the data collection used in the experiments and how we estimate the reliability of a metric.
  
  \subsection{Data Collection and Metric Evaluation}
  \label{sec:data_collection}
  

  To evaluate the correlation of reference-based metrics with human judgments, we collect a dataset as follows.
  We first crawled 5,000 queries from some Chinese social websites and the largest reference set for the obtained queries has 200 responses. Then, we generated 14 responses obtained from several widely used response generation models (described in Appendix~A). 
  We asked 5 annotators to assign each generated response with a score ranging from 1 to 5 respectively, and the final score of each response was obtained by averaging the five scores. Table~\ref{tab:data} shows the data statistics. See Appendix~A for a detailed description of our dataset.

  \begin{table}
    \centering
    \begin{tabular}{lr}
      \toprule


      \textbf{Raw Dataset}&\\
      \# Training Samples&4,000\\
       \# Validation Samples&500\\
        \# Test Samples&500\\
        \midrule
        \textbf{Each Sample}&\\
        \# Model Responses &14 \\ 
        \# Reference Responses & $\leq$ 200 \\ 
        \bottomrule
    \end{tabular}
    \caption{Our dataset statistics. Each sample is in the form (query,
    model responses, reference responses).}
    \label{tab:data}
  \end{table}
  
  To evaluate the performance of a metric, we leverage the Pearson Correlation. Given a sample consisting of a query $\bm{q}$, 14 generated responses $\bm{\mathcal{Y}}=\{\bm{y}_1,\cdots,\bm{y_{14}}\}$ with their human annotated scores $\hat{\bm{S}}=\{\hat{\bm{s}}_1,\cdots,\hat{\bm{s}}_{14}\}$ and a set of reference responses $\bm{\mathcal{R}}= \{\bm{r}_1,\cdots,\bm{r}_n\}$ ($n\leq200$), we first score each model response $\bm{y} \in \bm{\mathcal{Y}}$ with a certain metric, yielding a sequence of automatic evaluated scores $\bm{S}=\{s_1,\cdots,s_{14}\}$. With the automatic evaluated scores $\bm{S}$ and human annotated scores $\hat{\bm{S}}$ for the 14 generated responses, we can obtain the reliability score $c$ for each sample using the Pearson Correlation:
  \begin{equation}
      c = \mathrm{Pearson}(\bm{S},\hat{\bm{S}}).\label{eq:c}
  \end{equation}
 
  \subsection{Preliminary Experiments}
  \label{sec:pre_exp}
  Next, we conduct our preliminary experiments using the 500 samples in the test set and present our empirical observations regarding the two questions in beginning of Section~\ref{sec:assumption}. In the experiments in this section, the correlation of a certain metric is obtained by averaging the reliability scores over all 500 samples.  
  
\begin{figure}[t]
    \centering
    \includegraphics[width=0.9\linewidth]{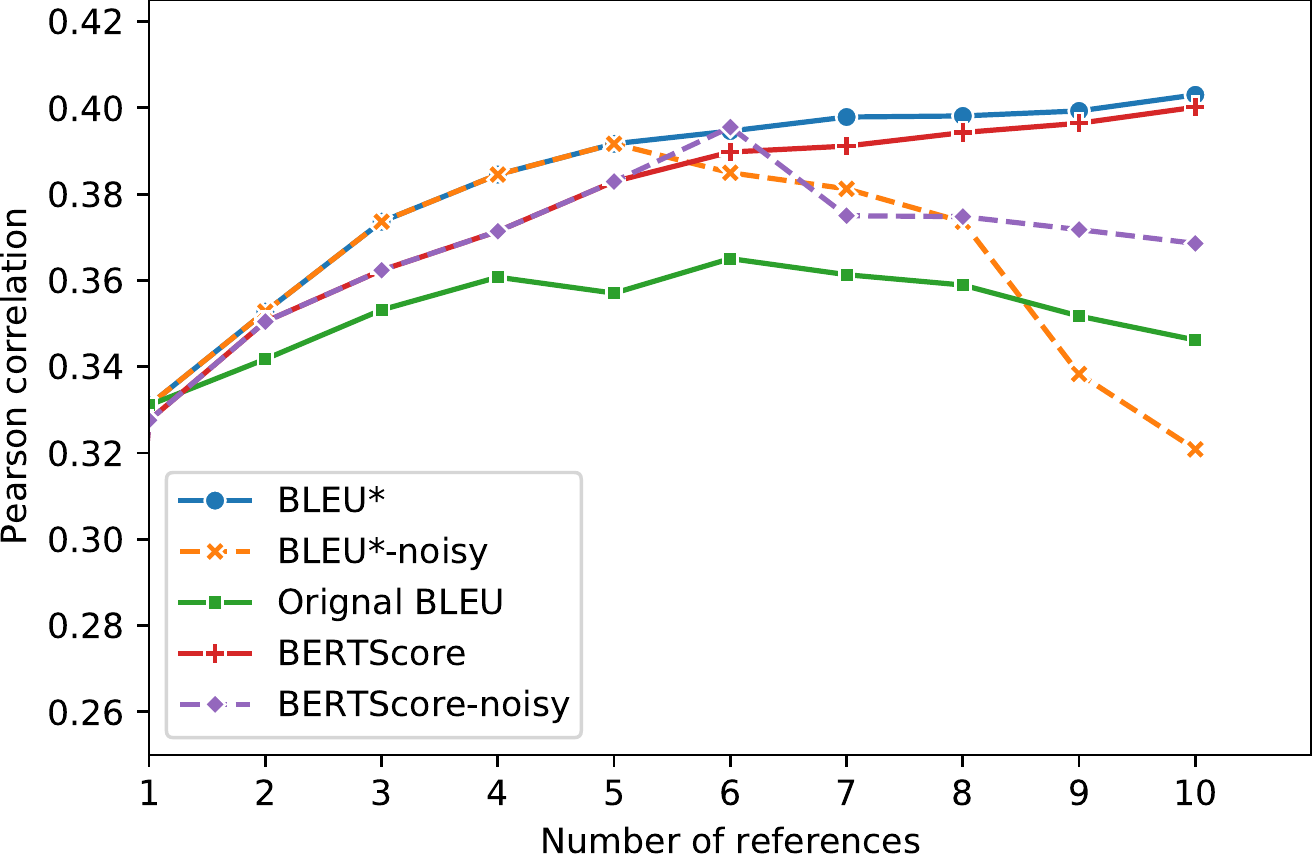}
    \caption{Pearson correlations of BLEU, BLEU* and BERTScore with human judgements obtained using different numbers of reference responses.}
    \label{fig:pre_exp}
  \end{figure}

  First, we would like to see how it affects the correlation of the metrics when high-quality responses are continuously added to the reference set.
  To ensure that each added reference response is of high quality, we also have human annotators assign a quality score to each crawled reference response, and then randomly select 10 high-quality reference responses (quality score over 4) for each sample in the test set. We sequentially add the 10 high-quality reference responses to the reference set initialized with an empty set. Each time a new reference response is added to the set, we calculate a reliability score of a certain metric using the updated set.
   Figure~\ref{fig:pre_exp} shows the evaluation results of the original BLEU, BLEU* and BERTScore. Noticeably, the correlation of the original BLEU does not improve as the number of high-quality sentences increases. The reason may be that the original BLEU is defined at the corpus-level and the n-gram precisions are sums over all corpus sentences, which means the newly added reference will cause the value to fluctuate.
  We, however, find that BLEU* and BERTScore are consistent with our assumption that the correlation of the metric improves as the number of high-quality responses in the set increases. Therefore, not all metrics meet our assumption, and we use BLEU* and the BERTScore for our following experiments. 

  Second, we test how low-quality responses would affect the correlation of the metric by adding noise into the reference set. In the noisy reference set, the first 5 responses in the reference set are added with the same high-quality responses as above, while the later 5 added responses are replaced with negative samples which are responses sampled from other queries. As shown in Figure~\ref{fig:pre_exp}, ``BLEU*-noisy'' and ``BERTScore-noisy'' denote the results of BLEU* and BERTScore using the noisy reference sets, respectively. As the number of negative samples in the set increases, the performance of the metrics starts to degrade. This further confirms our assumption that only adding more high-quality responses to the reference set will help the metric.

Based on the above analysis, we can see that for reference-based metrics satisfying our assumption, we can enhance its reliability by augmenting high-quality references to calculate the metric.
Next, a key question arises, how can we augment high-quality references? In the following, we propose a reliability prediction model to estimate the reliability of the reference set (Section~\ref{sec:approach}) and provide effective approaches to augment high-quality references with the help of the reliability prediction model (Section~\ref{sec:augment}).
  
  %
  
  


  \section{The Reliability Prediction Model}
  \label{sec:approach}
   In this section, we aim to estimate the reliability of a metric, when we use it to compute a performance score of an output response based on a given gold response set. 
  We formulate this task as a regression problem.
  Formally, given a query sentence $\bm{q}=\{q_1,\dots,q_M\}$ of length $M$, and a reference set $\mathcal{R} = \{\bm{r}_i\}_{i=1}^N$ with $N$ gold responses of the query, our goal is to learn a function $\bm{f}: (\bm{q},\mathcal{R}) \rightarrow c$ that predicts a reliability score $c$ that represents how reliable a metric is used to evaluate results of the input query using this reference set. 
  In this work, we consider using Pearson correlation ($c$) in Eq.~\ref{eq:c} between the human evaluation results and the metric scores from the given $\mathcal{R}$ as the reliability score of each $(\bm{q},\mathcal{R})$. 
   We introduce the reliability prediction model in this section, and discuss efficient methods to augment high-quality responses based on the trained prediction model in the next section.

  


  \begin{figure*}[htbp]
    \centering
    \includegraphics[width=0.9\linewidth]{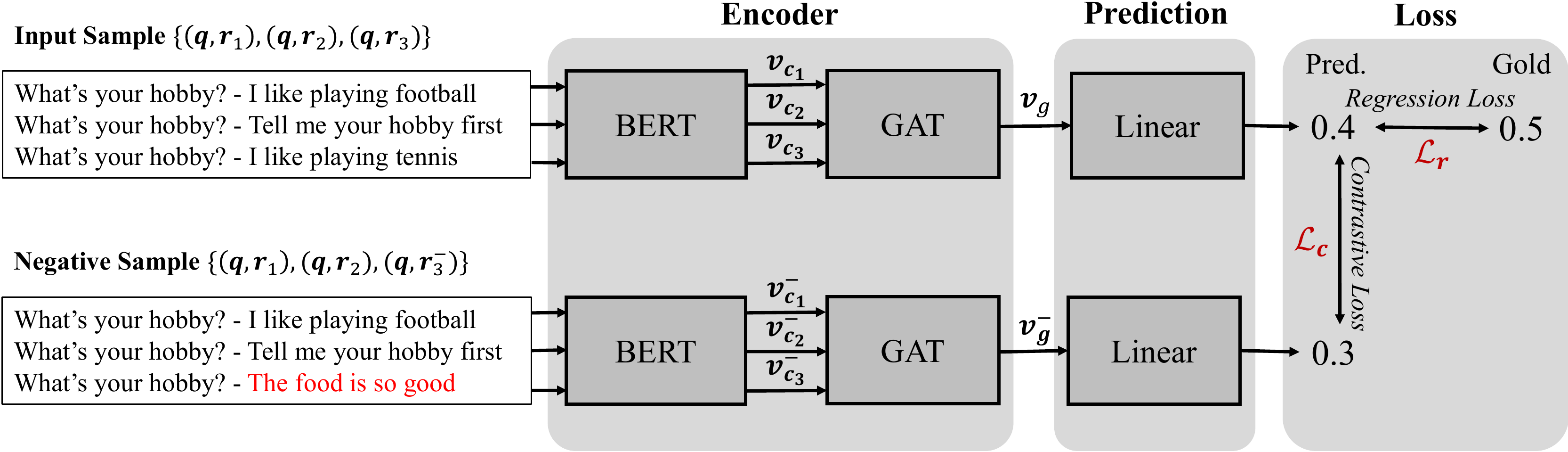}
    \caption{Our proposed prediction model. For each training sample $(\bm{q},\{\bm{r}\})$, we construct negative samples $(\bm{q},\{\bm{r}\}^-)$ according to our assumption and compute the contrastive loss. The model is trained with the combination of regression loss and contrastive loss.}
    \label{fig:arch}
  \end{figure*}
  
  Our learning framework is shown in Figure~\ref{fig:arch}, which contains two parts. We first design a prediction model to predict a reliability score for each $(\bm{q},\mathcal{R})$. Then, we construct negative samples and use contrastive learning to train the proposed prediction model.

 \subsection{Model Structure}
  \noindent
  \textbf{Input Representation}
  For a query $\bm q$ and each response $\bm{r}$ in the reference response set $\mathcal{R}$, we first concatenate them into one sequence. Then we can obtain $N$ query-response pairs.
  We leverage BERT~\citep{Devlin2019BERTPO} that learns contextualized representations of the query-response pairs: 
  \begin{equation}
    \!\!\! [\bm{v}_c,\bm{v}_{q_1},\cdots,\bm{v}_{r_1},\cdots,\bm{v}_{r_m}]=\mathrm{BERT}([\bm{q},\bm{r}])
  \end{equation}
  where $\bm{v}_c$ is the representation for the special token [CLS] in the BERT and we use its contextualized representation for each query-response pair. Then, we get $N$ query-response representations.

  \noindent
  \textbf{Graph Encoder:}
  In order to transform the dialogue context features into higher-level features that consider the intrinsic variance between reference responses in the reference set, we represent all query-response pairs in a fully-connected graph and each query-response pair is a node in the graph. Then we adopt a graph attention layer~\citep{Velickovic2018GraphAN} to obtain a better representation of each query-response pair: 
  \begin{equation}
   [\hat{\bm{v}}_{c_1},\ldots, \hat{\bm{v}}_{c_n}] =  \mathrm{GAT}([\bm{v}_{c_1},\ldots, \bm{v}_{c_n}]).
 \end{equation}
 The final representation $\bm{v}_g$ of the graph is computed using the max-pooling strategy.

\noindent
\textbf{Reliability Score Prediction:}
The reliability score is computed using a single-layer feedforward network coupled with a tanh activation function:
  \begin{equation}
      f(\bm{q},\mathcal{R}) = \mathrm{tanh}(\mathbf{W}\cdot \bm{v}_g +\mathbf{b})
  \end{equation}
  where $\mathbf{W}$ and $\mathbf{b}$ are trainable parameters.
  The model is trained to minimize the squared error between the model prediction and the gold correlation coefficient $c$ with L2-regularization:
  \begin{equation}
    \mathcal{L}_{r} = (f(\bm{q},\mathcal{R}) - c)^2 + \gamma\| \theta\|_2
  \end{equation}
  where $\gamma$ is a hyper-parameter and $\theta$ denotes the parameters of the model.
  
  \subsection{Data Augmentation}
  \label{sec:data_aug}
  To address the problem of limited data, we adopt the data augmentation method to improve the generalization of our models. Our input sample consists of a query $\bm{q}$ and a set of reference responses $\mathcal{R}$. We can expand the training data by using different combinations of reference responses, generating more different reference sets for each query. 
  %
   Since the number of all combinations is huge, we randomly sample some various combinations for each query. Given a set of reference responses $\bm{\mathcal{R}}= \{\bm{r}_1,\cdots,\bm{r}_n\}$ of $n$ elements ($n\leq200$), the set of $k$-combinations are denoted as $\bm{\mathcal{C}}_n^k$, where each combination has $k$ different elements. In our work, we use $k\in \{3,5,7,10\}$ and randomly sample 25 different combinations from each set $\bm{\mathcal{C}}_n^k$ ($k\in \{3,5,7,10\}$). We then expand the validation set in the same way.
   
   To evaluate the generalization performance of the model, the test set should be used not only to verify the effectiveness of the model on samples with $k \leq 10$, but also to test the performance of the model on samples with $k > 10$. Therefore, we use $k\in\{3,5,10,20,30,40\}$ to augment the test set. For each set $\bm{\mathcal{C}}_n^k$ ($k\in \{3,5,10,20,30,40\}$), we only randomly sample one combination. Finally, applying data augmentation as described above gives us a total of 400,000/4,000/3,000 augmented training/validation/test samples.
   
  \subsection{Contrastive Learning}
  
  Optimizing the simple regression objective above with the augmented training samples may not yield a robust performance. 
  To make the model capable to capture the differences between different reference set for the same query, 
  we construct multiple negative samples for each training sample $(\bm{q},\{\bm{r}\})$, and use contrastive learning to train the prediction model. Given a training sample ($\bm{q},\{\bm{r}\}$) we design three kinds of negative samples ($\bm{q},\{\bm{r}\}^{-}$):
  \begin{itemize}[wide=0\parindent,noitemsep, topsep=0pt]
    \item Remove one response $r$ from the existing reference set $\{\bm{r}\}$. This is based on the understanding in Section~\ref{sec:assumption} that a set of gold responses generally does not yield a higher correlation than a super set of it.
    We note that all samples with deteriorated correlation here, are treated as negative samples.
    \item Randomly select a response of any other query to add to the existing set. This is based on the understanding in Section~\ref{sec:assumption} that a noise response included in a gold reference set should deteriorate the correlation.
    \item Randomly select a response of any other query to replace a response in the existing set. The intuition is the same as the above one.
  \end{itemize}
  The contrastive loss function $\mathcal{L}_{c}$ with $T$ negative samples constructed is computed as:
  \begin{equation}
    \mathcal{L}_{c} = \frac{1}{T}\sum_{t} \mathrm{max}\{0,\Delta -f(\bm{q},\{\bm{r}\})+f(\bm{q},\{\bm{r}\}_t^{-})\},
  \end{equation}
  where $\Delta$ is a margin. The final loss can be computed as:
  \begin{equation}
    \mathcal{L} = \mathcal{L}_{r} + \mathcal{L}_{c}.
  \end{equation}
  
  
  \section{Augmenting High-quality References}
  \label{sec:augment}
  In this section, we describe how we can augment high-quality references based on the current gold reference set for a given query with the help of the proposed reliability prediction model. 
  Suppose a large set of reference candidates can be easily obtained. For example, more conversation data can be assessed to build a retrieval system to search for more references for the given query.
  We introduce two ways to handle reference candidates with different qualities.
  If the acquired reference candidates are considered reliable, we can adopt automatic annotation.
  If we are not certain about the relevance of the reference candidates, human annotators are needed for an interactive annotation.
  
  \subsection{Automatic Annotation}
  \label{sec:auto_anno}
  We assume the response candidates are all relevant to the given query. However, there may exist uninformative responses, such as the generic responses or those similar to the gold references.
  We now introduce how to use the predicted scores to automatically select out high-quality responses from the candidate set. 
  Each time a response is randomly picked to tentatively add into the gold reference set, the model will predict a reliability score for the augmented reference set. 
  If the reliability score of the augmented reference set improves, this picked response can be considered as a high-quality one and now confirmed to add to the gold reference set. Otherwise, we remove it from the gold reference set.
  We continue this process until all response candidates have been picked once, and return the final augmented gold reference set.  
  
  \subsection{Interactive Annotation}
  If the response candidates are of uneven relevance to the query,
   we need to hire annotators to manually check and edit the response candidates. 
   We design an interactive annotation strategy to allow the model to assist annotators.
   For each selected response candidate, the annotator mainly executes the following three steps:
   \begin{enumerate}[wide=0\parindent,noitemsep, topsep=0pt]
      \item  If he/she considers the  response candidate is relevant to the given query and the reliability score of the set improves with the current candidate response, he/she can retain this response directly;
    \item If he/she considers the response candidate is not relevant enough to the given query but the reliability score of the set shows improvement with the current candidate response, he/she needs to edit this response and check whether the reliability score of the set increases with the edited response;
    \item  If the reliability score of the set does not improve with the current candidate response, no matter whether the candidate response is relevant or not, he/she still needs to edit this response and check whether the reliability score of the set increases with the edited response;

   \end{enumerate}
       In both Step2 and Step3, we allow the interactive annotation with a maximum number of attempts. Otherwise, we abandon this response and continue to the next response candidate.
       This process can help annotators avoid writing responses with unsatisfactory quality.

  \section{Experiments}
    \label{sec:main_exp}
  \subsection{Setup}

  In our experiments, we use the quality scores obtained by BLEU* and BERTScore to compute the Person correlations with human judgments respectively.
  The reliability models introduced in Section~\ref{sec:approach} trained with BLEU and BERTScore are referred as \textbf{REAM$\sharp$(BLEU*)} and \textbf{REAM$\sharp$(BS)} respectively. The augmented dataset mentioned in Section~\ref{sec:data_aug} is used for the corresponding model training and testing. 
  To show its effectiveness on automatic annotation and interactive annotation, we use the 500 test samples in the raw dataset introduced in Section~\ref{sec:data_collection}. 
  See Appendix~B for implementation and training details.

\subsection{Results on Reliability Prediction Models}
\begin{table}
  \centering
  \begin{tabular}{cccc}
    \toprule
     \textbf{\# Refs}  & \textbf{MSE} & \textbf{Pred.} & \textbf{Gold} \\
     \midrule
     \multicolumn{4}{l}{\textbf{REAM$\sharp$(BLEU*)}} \\
    3 & 0.006 \small{$\pm $0.005} & 0.333 & 0.340 \\ 
    5 &  0.006 \small{$\pm $0.006} & 0.374 & 0.379 \\ 
    10 &  0.006 \small{$\pm $0.005} & 0.399 & 0.402 \\ 
    20 & 0.006 \small{$\pm $0.005} & 0.412 & 0.409 \\ 
    30 & 0.007 \small{$\pm $0.006} & 0.413 & 0.417\\ 
    40 & 0.006 \small{$\pm $0.006} &  0.422 & 0.425 \\ 
      \midrule
      \multicolumn{4}{l}{\textbf{REAM$\sharp$(BS)}} \\
      3 & 0.006 \small{$\pm $0.006}  & 0.356 & 0.355 \\ 
      5 & 0.006 \small{$\pm $0.005} & 0.367 & 0.375 \\ 
      10 & 0.006 \small{$\pm $0.006} & 0.393 & 0.404  \\ 
      20 & 0.006 \small{$\pm $0.005} & 0.397 & 0.408 \\ 
      30 &  0.006 \small{$\pm $0.005}  & 0.406 & 0.417 \\ 
      40 & 0.007 \small{$\pm $0.006} & 0.417 & 0.431 \\ 
      \bottomrule
  \end{tabular}
  \caption{Results of the \textbf{REAM$\sharp$(BLEU*)} model and the \textbf{REAM$\sharp$(BS)} model on multiple test sets consisting of different number of reference responses. 
  }
  \label{tab:model}
\end{table}
In Table~\ref{tab:model}, we report MSE, the averaged predicted reliability scores and averaged gold reliability scores. Standard deviations are also provided.
We can see that the MSE of both prediction models are kept at a low level on different test sets. The difference between the average predicted reliability score and the average gold scores is at most 0.014. Also, the models have good stability and rarely show extreme cases, as shown by their standard deviations of less than 0.01. Our proposed models also have good generalization performance in terms of different sizes of the reference set. The number of references of training samples in the training set is at most 10, but the two models still have a small test error on the test set with more than 10 references, which is comparable to the test set with less than 10 references. 

\begin{figure}[t]
  \centering
  \includegraphics[width=0.9\linewidth]{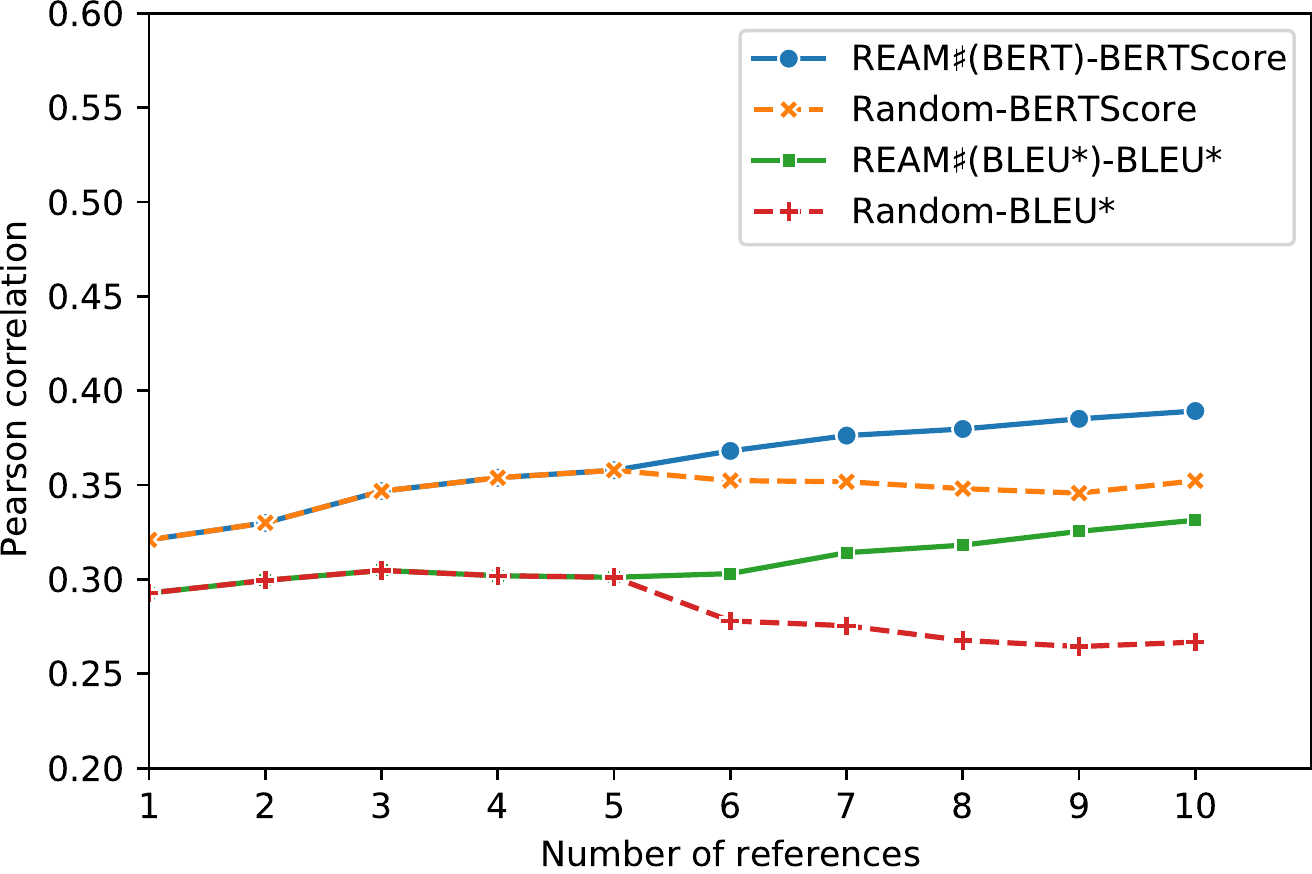}
  \caption{Pearson correlations of BLEU* and BERTScore on multiple reference sets obtained by different strategies. ``REAM$\sharp$(BS)-BERTScore'' means using the \textbf{REAM$\sharp$(BS)} to augment the reference set and testing it with \textbf{BERTScore}. ``REAM(BLEU*)-BLEU*'' means using the \textbf{REAM$\sharp$(BLEU*)} to augment the reference set and testing it with \textbf{BLEU*}. We also add a random strategy for comparison.}
  \label{fig:auto_results}
\end{figure}

\subsection{Results on Automatic Annotation}
Once we have a reliable model, the next is to use the model to help us collect high-quality references. When enough candidate responses are available, we can directly use the model to identify which responses are helpful to improve the reliability of the set. The experiment on automatic annotation is conducted with the raw test set introduced in Section~\ref{sec:data_collection}.
For each sample in the test set, we initialize the reference set with a randomly sampled crawled reference response, then add the remaining reference responses (as mentioned before, each sample has at most 200 reference responses) to its reference set one by one. Each time a new response is added, the model predicts a reliability score. The augmentation method follows the strategy mentioned in Sec.~\ref{sec:auto_anno}.
We selected 10 responses for each sample. 

Figure~\ref{fig:auto_results} shows the Pearson correlations of BLEU* and BERTScore on multiple reference sets obtained by different strategies. ``REAM$\sharp$(BS)-BERTScore'' means using the \textbf{REAM$\sharp$(BS)} model to augment the reference set and testing it with \textbf{BERTScore}. Similarly, ``REAM(BLEU*)-BLEU*'' means using the \textbf{REAM$\sharp$(BLEU*)} model to augment the reference set and testing it with \textbf{BLEU*}. We also use a random strategy for comparison (orange and red), where the first 5 responses are the same as those previously selected using the models, and the last 5 responses are obtained by randomly sampling.
As shown in the figure, the reliability of the reference sets constructed using our proposed method tends to increase steadily as more selected responses are added to the set, while the reliability of the reference sets constructed using the random strategy appears to be unstable. We also test the transferability of the model and find that the model trained with one metric also yields reliable performance when tested with other metrics. The results are shown in Appendix~E.

The final results are shown in Table~\ref{tab:final} which reports Pearson and Kendall correlations of BLEU* and BERTScore calculated using three reference sets. ``Raw'' denotes the initial reference set containing the first selected reference response. The remaining 9 selected responses for each sample are then used to augment the ``Raw'' set. ``Aug-REAM$\sharp$(BS)'' and ``Aug-REAM$\sharp$(BLEU*)'' are the reference sets augmented using the \textbf{REAM$\sharp$(BS)} and \textbf{REAM$\sharp$(BLEU*)}, respectively. ``Mix'' is the union of the two sets. 
As can be seen, the performances of the two metrics both improve using the augmented reference set. 
When combining the two augmented reference sets, both Pearson and Kendall correlations of BERTScore improve, while for BLEU*, the Pearson and the Kendall correlations dropped slightly. This indicates that BERTScore is better at capturing semantics than BLEU* and is able to select the most adequate reference response from multiple augmented sets for evaluation.

\subsection{Results on Interactive Annotation}
\label{sec:interactive}

In the following, we perform experiments to augment high-quality references of 30 queries randomly selected in the test set in Section~\ref{sec:data_collection} using interactive annotation introduced in Section~\ref{sec:interactive}. 
For each test query, we first use ElasticSearch~\footnote{Elasticsearch is a search engine based on the Lucene library. \url{https://www.elastic.co}} to retrieve 100 candidate responses from a database with 200 million (query, response) pairs with Jaccard similarity for each query,
and display them to each annotator.
We recruit six annotators and divide them into two groups. 
The reliability scores predicted by the \textbf{REAM$\sharp$(BERT)} model will reveal to 
annotators in the first group (``Human-REAM$\sharp$(BS)'') for interactive annotation but not annotators in the second group (``Human'').
Annotators in ``Human'' perform non-interactive annotations.
They are required to rewrite their considered unsuitable responses directly without the predicted scores as a reference. 
For each round of annotation,
we pick one identical candidate response to all annotators.
%
%
\begin{table}
  \centering
  \begin{tabular}{lcc}
    \toprule
    &\textbf{Pearson}&\textbf{Kendall}\\
    \midrule
      \textbf{BLEU*}&&\\ 
      \ \ Raw & 0.246&0.165 \\ 
      \ \ Aug-REAM$\sharp$(BLEU*)&\textbf{0.331}&\textbf{0.269} \\ 
      \ \ Aug-REAM$\sharp$(BS)&0.315&0.261 \\ 
      \ \ Mix&0.329&0.266 \\ 
      \midrule
      \textbf{BERTScore} && \\ 
      \ \ Raw&0.288&0.209 \\ 
      \ \ Aug-REAM$\sharp$(BLEU*)&0.380&0.311 \\ 
      \ \ Aug-REAM$\sharp$(BS)&0.389&0.318 \\ 
      \ \ Mix&\textbf{0.393}&\textbf{0.319} \\ 
      \bottomrule
  \end{tabular}
  \caption{Pearson and Kendall correlations of BLEU and BERTScore calculated using different constructed reference sets. ``Raw'' denotes the reference set consisting of 1 response.``Aug-REAM$\sharp$(BS)'' and ``Aug-REAM$\sharp$(BLEU*)'' are the reference sets (10 reference responses) augmented using the \textbf{REAM$\sharp$(BS)} and \textbf{REAM$\sharp$(BLEU*)}, respectively. ``Mix'' is the union of the two sets.}
  \label{tab:final}
\end{table}

\begin{figure}[htbp]
  \centering
    \includegraphics[width=0.9\linewidth]{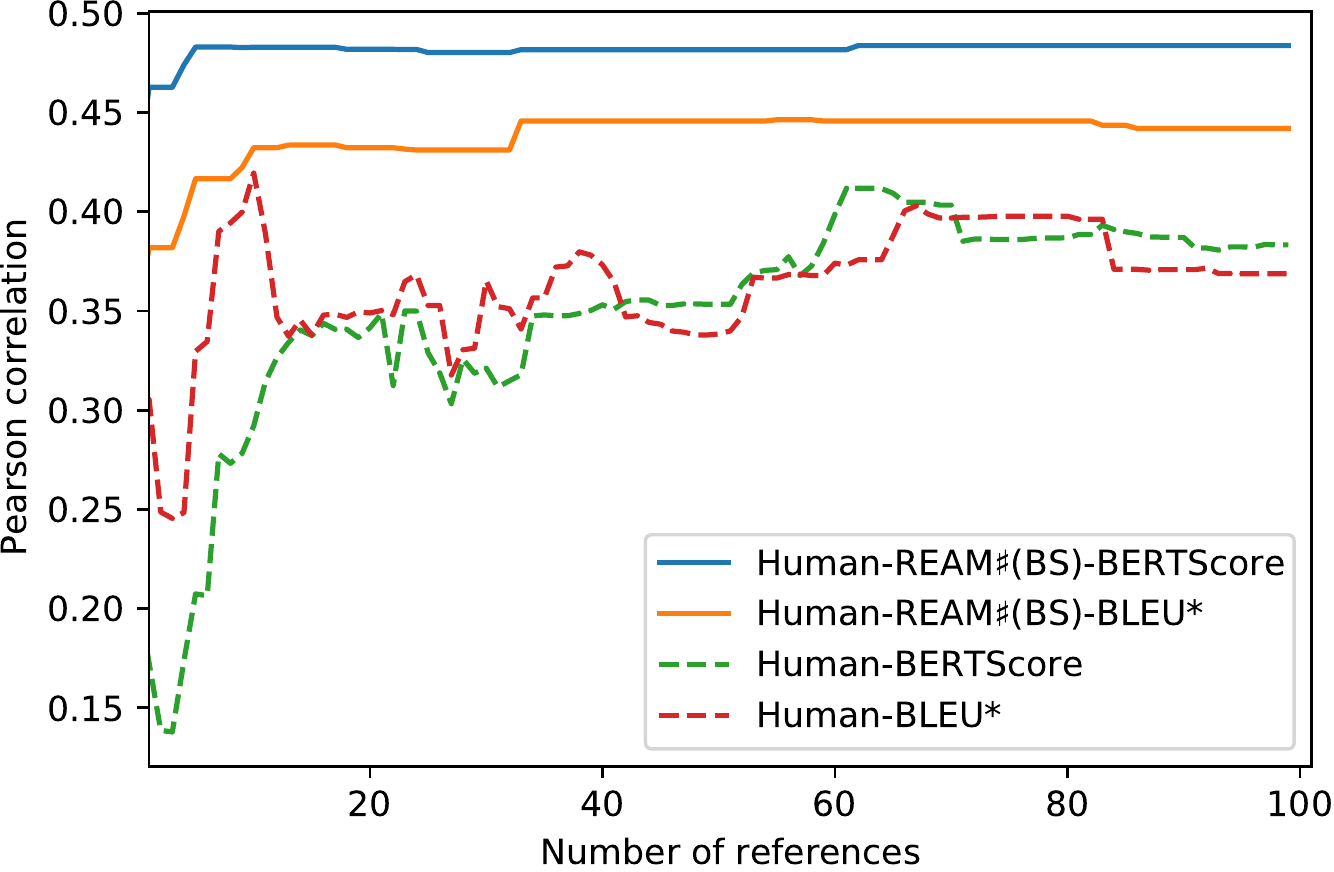}
    \caption{Pearson correlations of the reference sets constructed with/without the \textbf{REAM$\sharp$(BS)} model in interactive annotations, respectively. ``Human-REAM$\sharp$(BS)'' denotes the annotator with the model assistance and ``Human'' is the annotator without the model assistance.}
    \label{fig:human_bert}
\end{figure}

Figure~\ref{fig:human_bert} shows the Pearson correlations of human evaluation results and BERTScore with the augmented reference sets after each round of annotation by one annotator in ``Human-REAM$\sharp$(BS)''
and another annotator in ``Human''.
See Appendix~D for the results of all six annotators. 
We can see that using the augmented response set from annotators in the first group has already reached a relatively high Pearson correlation with only a few annotation rounds. When the number of responses increases to a certain level, the reliability score of the metric hits a bottleneck and rises more slowly. 
However, the correlation results using the augmented response set from annotators in the second group are not stable. Though the overall trend is increasing, the final obtained reference set has even worse correlations than a much small augmented set from the first group. This shows that our interactive annotation strategy is effective to help annotators avoid writing responses with unsatisfactory quality. 

  \section{Related Work}
  \label{sec:related_work}
  Automatic evaluation is crucial to the research of open-domain dialog systems~\citep{Li2016MutualIA,Li2017AdversarialLF,Gao2019GeneratingMD,Venkatesh2018OnEA,chan2021enhancing,xiang2021assessing}. 
  Existing metrics can be broadly categorized into reference-based and reference-free metrics~\citep{wang2021a}. 
  In this work, we focus on reference-based metrics, which usually measure how similar a generated response is to the reference response.


  The most commonly used reference-based metrics for dialog systems were originally proposed for machine translation. They typically count the amount of word-overlap between the generated response and the reference response. BLEU~\citep{Papineni2002BleuAM} is the most widely used metric in machine translation that calculates the geometric mean of the precision for $n$-gram. 
  Other related word-overlap metrics such as NIST~\citep{Lin2004AutomaticEO}, METEOR~\citep{Banerjee2005METEORAA} and ROUGE~\citep{Lin2004ROUGEAP} also have been used for dialogue evaluation.

  Instead of using word-overlap based metrics, embedding-based metrics are adopted to consider the semantic meaning of a word as defined by a distributed representation. They typically compute the similarity between the generated response and reference response using approximated sentence-level representations.
  The most commonly used word embedding based metrics use a heuristic to combine the vector representation of the individual word in the sentence. For example, Embedding Average~\citep{Foltz1998TheMO, Mitchell2008VectorbasedMO},
   Vector Extrema~\citep{Forgues2014BootstrappingDS}, 
   and Greedy Matching~\citep{Rus2012ACO}.
    \citet{Zhang2020BERTScoreET} introduce a better embedding-based metric BERTScore that computes token similarity using contextual embeddings that capture the specific use of a word in a sentence. 
  
  A few reference-based metrics for dialog systems are learnable functions. 
  ADEM~\citep{Lowe2017TowardsAA} which is based on neural networks is trained to predict a score of a response given its query and a reference response. RUBER~\citep{Tao2018RUBERAU}  evaluates responses with a blending of scores from the referenced and unreferenced metrics. 
  RUBER is learnable, but its training does not require human annotation scores.

  As discussed from the very beginning of our work, all the above work focus on designing a better metric. However, the reliability of the reference set is also a key to improve the correlation of reference-based metrics, but not investigated in detail in previous work. Therefore, we believe our work can fill this vacancy and provide a new direction  to improve reference-based metrics for open-domain dialogue generation.


  \section{Conclusions and Future Work}
  \label{sec:conclusion}
  In this paper, we first clarify an assumption on existing reference-based metrics that if more high-quality reference responses are added to the reference set, it should have a higher correlation with human judgment. For metrics satisfying this assumption, we present \textbf{REAM}$\sharp$, an enhancement approach to improve their reliability. In our approach, a reliability prediction model is trained to estimate the reliability of the reference set and we explore both automatic and interactive ways to augment high-quality references with the help of the reliability prediction model. Experiments show that our approach can efficiently help augment high-quality references and the correlations of reference-based metrics improve when using the augmented reference sets to evaluate dialog responses.
  %
  %
  Our work currently focuses on open-domain dialog systems as a starting point. However, the \textbf{REAM}$\sharp$ framework can be extended naturally to other open-ended text generation tasks such as story generation and question generation.

   \section{Ethical Considerations}
   The dataset used in our work are crawled from several Chinese social media websites, BaiduTieba, Douban, Weibo and Zhihu.  We purposefully avoid deanonymization techniques based on exploiting software vulnerabilities and our approach that involves human participation in rewriting responses gives us no access to any personally identifiable information.
  
  \bibliographystyle{acl_natbib}
  \bibliography{main}

  \newpage
  \appendix 

  \section{Dataset}
  \label{sec:app_data}
  The dataset used in our work are crawled from several Chinese social media websites, BaiduTieba, Douban, Weibo and Zhihu. We purposefully avoid deanonymization techniques based on exploiting software vulnerabilities and our approach that involves human participation in rewriting responses gives us no access to any personally identifiable information. We crawled 5,000 queries and each query has at most 200 ref-erence responses. Then, we collected 14 machine-generated responses obtained from several widely used responses generation models:
  \begin{itemize}
    \item LSTM-S2S-BS: a LSTM Seq2Seq model that generates responses with beam search.
    \item LSTM-S2S-Sampling: a LSTM Seq2Seq model that generates responses with top k sampling.
    \item LSTM-S2S-MMI: a LSTM Seq2Seq model with a Maximum Mutual Information-based decoding strategy.
    \item Fconv-S2S-BS: a convolutional Seq2Seq model~\citep{Gehring2017ConvolutionalST} that generates responses with beam search. 
    \item Fconv-S2S-Sampling: a convolutional Seq2Seq model~\citep{Gehring2017ConvolutionalST} that generates responses with top k sampling. 
    \item Fconv-S2S-DBS: a convolutional Seq2Seq model~\citep{Gehring2017ConvolutionalST} that generates responses with diverse beam search~\citep{Li2016ASF}.
    \item Transformer-S2S-BS: a Transformer Seq2Seq model that generates responses with beam search. 
    \item Transformer-S2S-Sampling: a Transformer Seq2Seq model that generates responses with top k sampling. 
    \item Transformer-S2S-DBS: a Transformer Seq2Seq model that generates responses with diverse beam search~\citep{Li2016ASF}.
    \item GPT2-BS: a GPT2~\citep{Radford2019LanguageMA} model that generates responses with beam search.
    \item GPT2-TopK Sampling (k=20): a GPT2~\citep{Radford2019LanguageMA} model that generates responses with top 20 sampling.
    \item GPT2-TopK Sampling (k=10): a GPT2~\citep{Radford2019LanguageMA} model that generates responses with top 10 sampling.
  \end{itemize}

  The GPT2 model are trained on a 200 million chinese dataset crawled also from the Chinese social media websites (BaiduTieba, Douban, Weibo and Zhihu). The LSTM-S2S, Fconv-S2S and Transformer-S2S models are trained on a benchmark dataset with 7M query-response pairs proposed by ~\citet{Liu2018TowardsLG}.

  \section{Implementation and Training Details}
  \label{sec:app_train}
    We leverage ``bert-as-service'' (\url{https://github.com/hanxiao/bert-as-service}), an open-source system that uses BERT as a sentence encoder and hosts it as a service to map sentences into fixed-length representations. In our work, we use the character-level BERT pre-trained in Chinese. Our reliability prediction model is implemented using ``PyTorch Geometric'' which is a geometric deep learning extension library for PyTorch. We use one layer. For graph encoding, we employ a one-layer graph attention network with an input size of 768. The input size of the prediction linear layer also is 768. The model is trained using Adam optimizer with a learning rate of 0.0005. The batch size is 128.

    For computing BLEU, we use the Python NLTK library. For computing BERTScore, we use the implementation provided by~\citet{Zhang2020BERTScoreET} at \url{https://github.com/Tiiiger/bert_score}.
  \section{Case Study}
\begin{figure*}[htbp]
    \centering
  
    \includegraphics[width=0.9\linewidth]{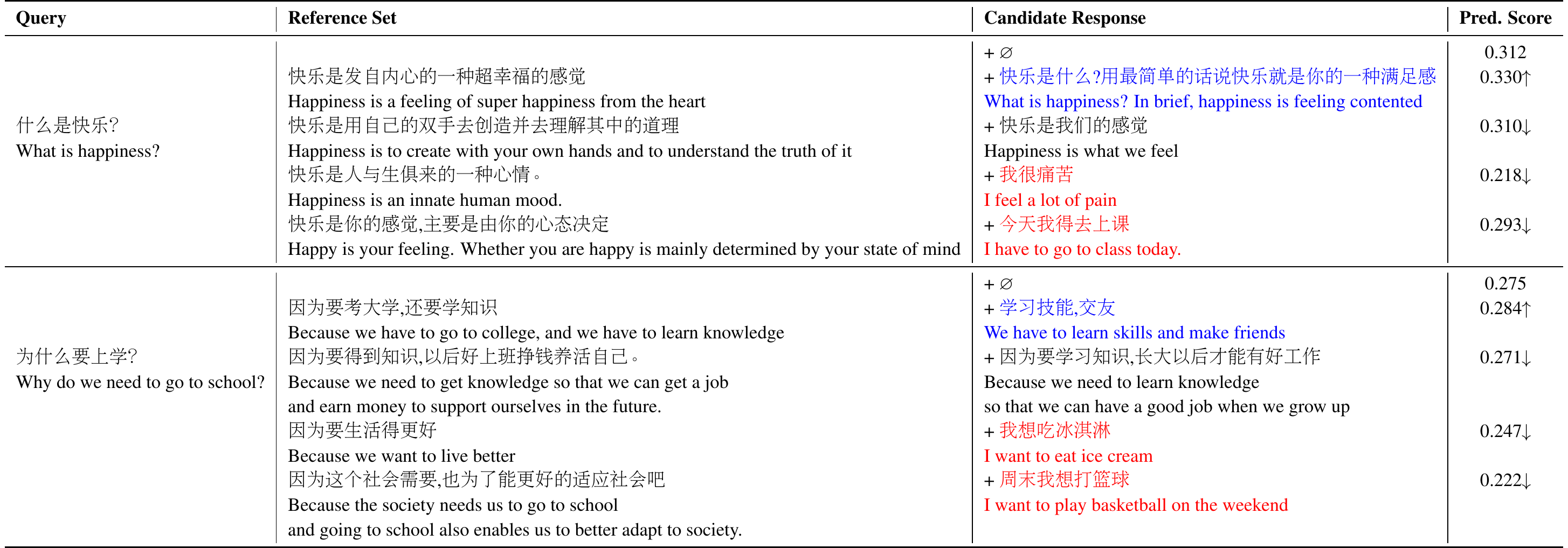}
    \caption{Two examples of using our \textbf{REAM$\sharp$(BS)} model to predict reliability scores for reference sets. The sentences with blue text are high-quality and diverse responses. Sentences without colour are responses that are similar in meaning to the responses already in the set. Red sentences are responses that are completely irrelevant for the queries.}
    \label{fig:case}

  \end{figure*}

  Figure~\ref{fig:case} shows two examples of using our \textbf{REAM$\sharp$(BS)} model to predict reliability scores for reference sets. Given a query and a reference set consisting of four reference responses, we first predict a reliability score for this set (e.g 0.312 and 0.275). We prepared four different candidate responses for each sample. The sentences with blue text are high-quality and diverse responses. Sentences without color are responses that are similar in meaning to the responses already in the set. Red sentences are responses that are completely irrelevant to the queries. We add each of the four candidate responses to the set and predict the reliability score for the new set. As shown in the figure, when the high-quality responses are added to the set, the model predicts a higher reliability score than the original reliability score. When responses with repeated semantics are added to the set, the model predicts a slightly lower reliability score compared to the original one. When poor quality responses are added to the set, the reliability scores predicted by the model drop sharply. It can be seen that our model can effectively evaluate the impact of responses with different qualities on the reference set.


  \section{More Results on Interactive Annotation}
  \label{sec:app_human}
\begin{figure*}[htbp]
  \centering
  \subfigure{
  \begin{minipage}[t]{0.33\linewidth}
  \centering
  \includegraphics[width=0.9\linewidth]{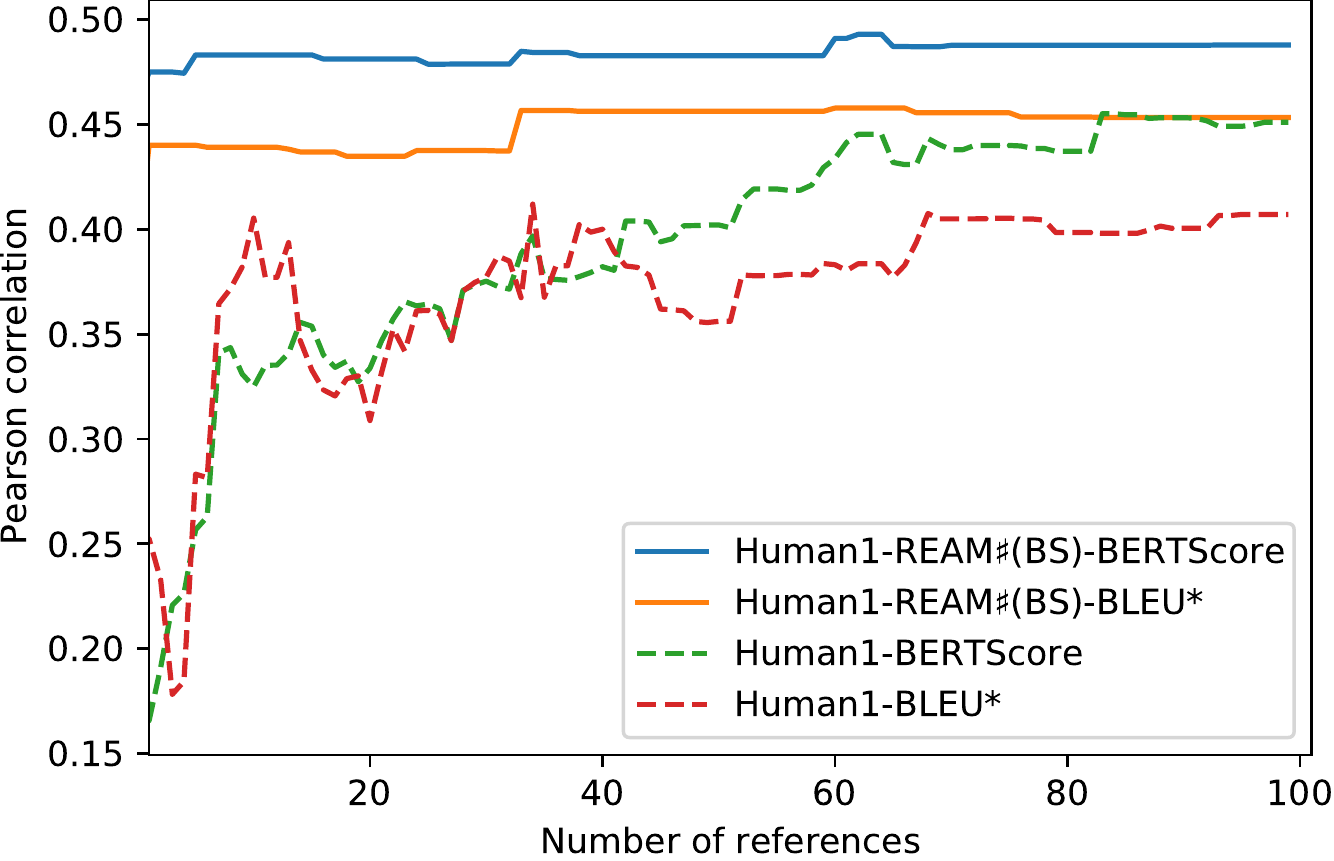}
  \end{minipage}%
  }%
  \subfigure{
  \begin{minipage}[t]{0.33\linewidth}
  \centering
  \includegraphics[width=0.9\linewidth]{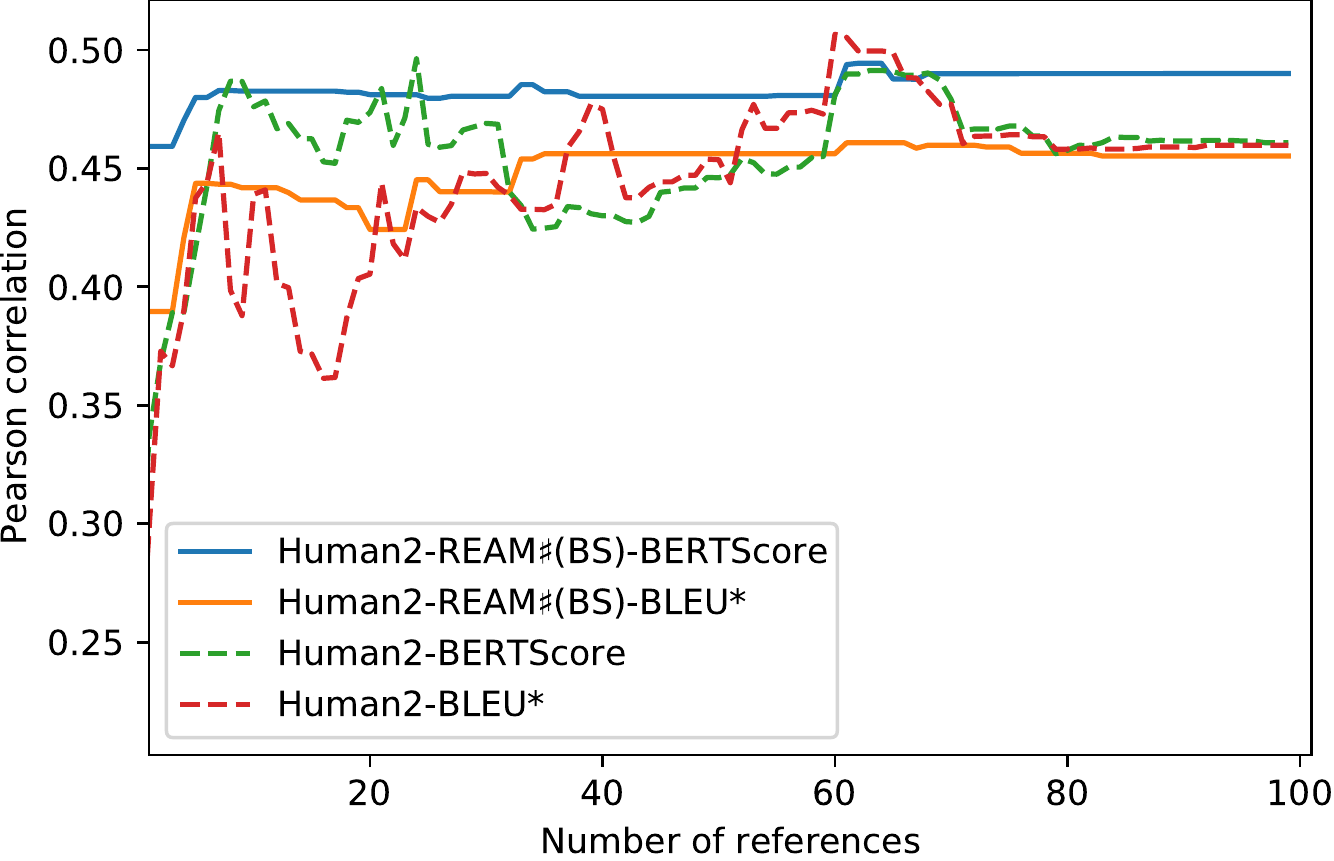}
  \end{minipage}%
  }%
  \subfigure{
  \begin{minipage}[t]{0.33\linewidth}
  \centering
  \includegraphics[width=0.9\linewidth]{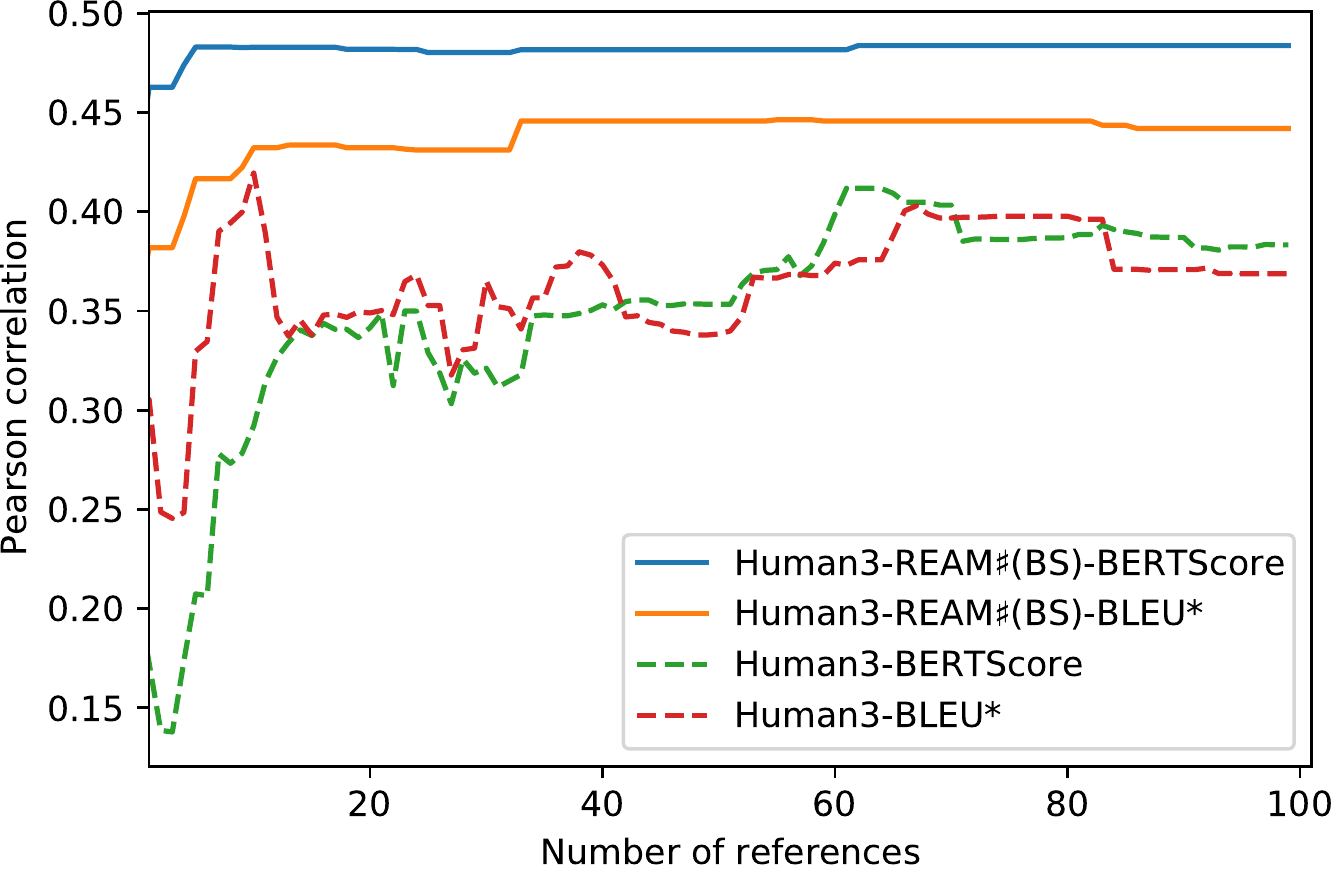}
  \end{minipage}%
  }%
  \caption{Pearson correlations of the reference sets constructed with and without the \textbf{REAM$\sharp$(BS)} model, respectively. ``Human-REAM$\sharp$(BS)'' denotes the annotator with the model assistance and ``Human'' is the annotator without the model assistance.}
  \label{fig:all_human}
  
\end{figure*}

  Figure~\ref{fig:all_human} shows the results of all six annotators.
  We can see that using the augmented response set from annotators in the first group has already reached a relatively high Pearson correlation with only a few annotation rounds. When the number of responses increases to a certain level, the reliability score of the metric hits a bottleneck and rises more slowly. 
  However, the correlation results using the augmented response set from annotators in the second group are not stable. Though the overall trend is increasing, the final obtained reference set has even worse correlations than a much small augmented set from the first group. This shows that our interactive annotation strategy is effective to help annotators avoid writing responses with unsatisfactory quality. 
  
\section{Transferbility}
\label{sec:transfer}
\begin{figure}[t]
  \centering
  \includegraphics[width=0.9\linewidth]{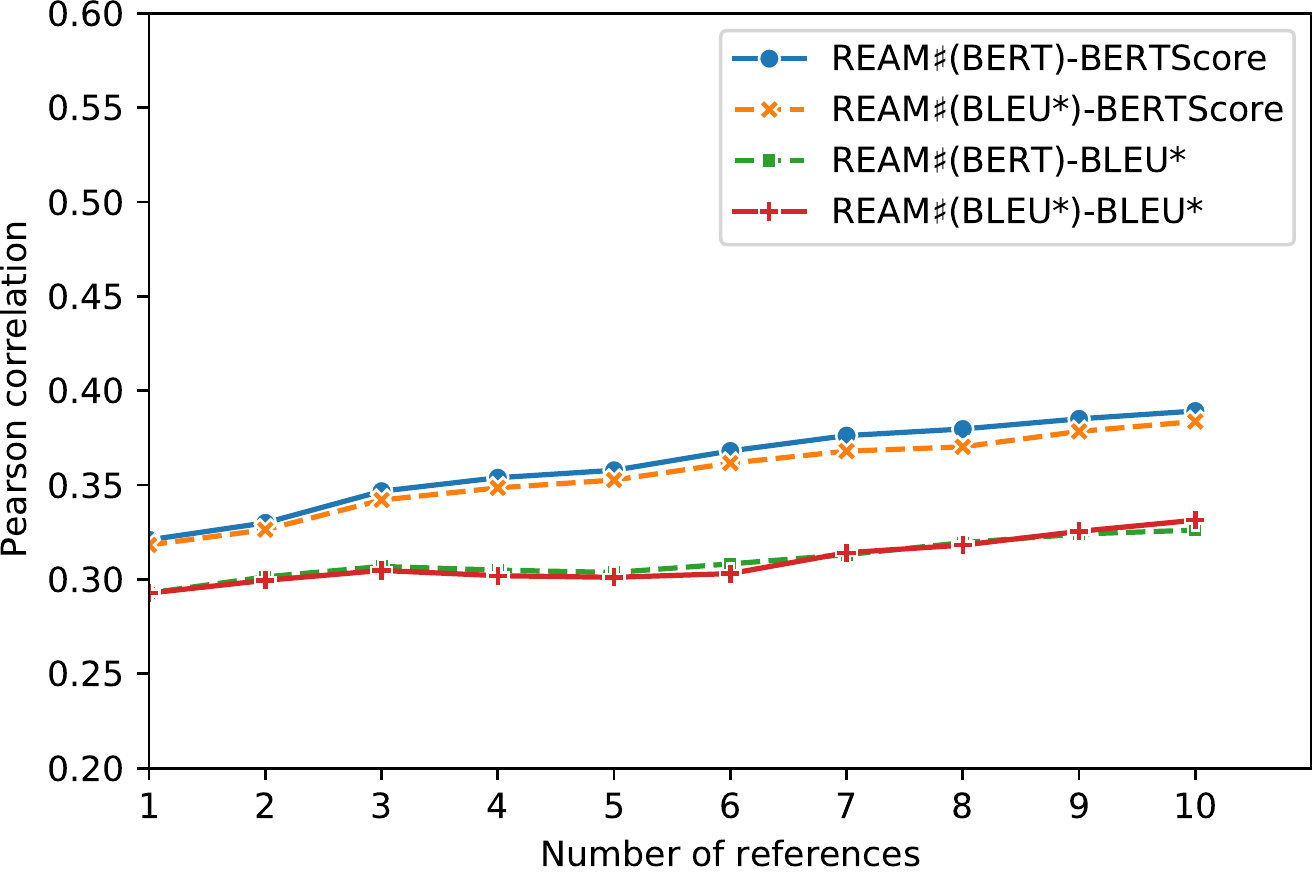}
  \caption{Pearson correlations of the two models, \textbf{REAM$\sharp$(BS)} and \textbf{REAM$\sharp$(BLEU*)} tested using BLEU* and BERTScore, respectively.}
  \label{fig:transfer}
\end{figure}

We would like to see if the model trained with one metric also yields reliable performance when tested with other metrics. Figure~\ref{fig:transfer} shows the Pearson correlations of the two models, \textbf{REAM$\sharp$(BS)} and \textbf{REAM$\sharp$(BLEU*)} tested using BLEU* and BERTScore, respectively. From the figure, we can see that whether tested with BLEU* or BERTScore, the difference in performance between \textbf{REAM$\sharp$(BS)} and \textbf{REAM$\sharp$(BLEU*)} is very small. We also notice that the difference in performance tested with BLEU* between the two models is somewhat larger than that tested with BERTScore when the number of references is large. This may be because BLEU* does not utilize the semantic information compared to BERTScore, which leads to some high-quality responses in the reference set being ignored. Therefore, our enhancement approach is more effective as the metric's ability to capture semantic similarity increases.

\end{document}